%% file: preprint.tex
\documentclass[lettersize,journal]{IEEEtran}

\usepackage{graphicx}
\usepackage{amsmath, amsfonts}
\usepackage{amssymb}
\usepackage{mathtools}
\usepackage{amsthm}
\usepackage{multirow}
\usepackage{bm}
\usepackage{utfsym}
\usepackage{color}
\usepackage{colortbl}
\usepackage[table]{xcolor}
\usepackage{array}
\usepackage{textcomp}
\usepackage{stfloats}
\usepackage{url}
\usepackage{verbatim}
\usepackage{cite}
\usepackage{arydshln}
\usepackage[caption=false,font=normalsize,labelfont=sf,textfont=sf]{subfig}
\usepackage{algorithm}
\usepackage{algorithmic}

\definecolor{green}{HTML}{81B996}
\definecolor{highgreen}{cmyk}{0.9, 0, 1, 0.3}
\definecolor{orange}{HTML}{F2AA84}
\definecolor{blue}{HTML}{4E95D9}
\definecolor{red}{HTML}{BE2F21}
\definecolor{gray10}{gray}{0.9}
\definecolor{gray20}{gray}{0.8}
\definecolor{highlightcyan}{cmyk}{0.1,0,0,0}

\newlength\thickarrayrulewidth
\makeatletter
\newcommand{\thickhline}{%
    \noalign {\ifnum 0=`}\fi \hrule height 1pt
    \futurelet \reserved@a \@xhline
}
\makeatother

\theoremstyle{plain}
\newtheorem{theorem}{Theorem}[section]
\newtheorem{proposition}[theorem]{Proposition}

\newtheorem{definition}[theorem]{Definition}

\newtheorem{remark}[theorem]{Remark}

\newcommand{\partialgray}{\rowcolor{gray10}\cellcolor{white}}
\begin{document}

\title{Universal Adversarial Attacks against Closed-Source MLLMs via Target-View Routed Meta Optimization}

\author{Hui Lu, Yi Yu, Yiming Yang, Chenyu Yi, Xueyi Ke, Qixing Zhang, Bingquan Shen, Alex Kot,~\IEEEmembership{Life Fellow,~IEEE,} \\
Xudong Jiang,~\IEEEmembership{Fellow,~IEEE}
\IEEEcompsocitemizethanks{
        \IEEEcompsocthanksitem Hui Lu is with the Rapid-Rich Object Search Lab, Interdisciplinary Graduate Programme, Nanyang Technological University, Singapore, (e-mail: hui007@e.ntu.edu.sg).
        \IEEEcompsocthanksitem Yi Yu, Chenyu Yi, Xueyi Ke, Alex C. Kot, and Xudong Jiang are with the School of Electrical and Electronic Engineering, Nanyang Technological University, Singapore, (e-mail: \{yuyi0010, cyyi, xueyi.ke, eackot, exdjiang\}@ntu.edu.sg).
  \IEEEcompsocthanksitem Yiming Yang and Qixing Zhang are with the College of Computing and Data Science, Nanyang Technological University, Singapore, (e-mail: \{yiming014, qixin.zhang\}@ntu.edu.sg).
\IEEEcompsocthanksitem Bingquan Shen is with the DSO National Laboratories, Singapore, (email: sbingqua@dso.org.sg)
}
}

\markboth{Journal of \LaTeX\ Class Files,~Vol.~14, No.~8, August~2021}%
{Shell \MakeLowercase{\textit{et al.}}: A Sample Article Using IEEEtran.cls for IEEE Journals}

\maketitle

\let\origTwoColumn\twocolumn
\renewcommand{\twocolumn}[1][]{}

\begin{abstract}

Targeted adversarial attacks on closed-source multimodal large language models (MLLMs) have been increasingly explored under black-box transfer, yet prior methods are predominantly \emph{sample-specific} and offer limited reusability across inputs. We instead study a more stringent setting, Universal Targeted Transferable Adversarial Attacks (UniTTAA), where a single perturbation must consistently steer arbitrary inputs toward a specified target across unknown commercial MLLMs. Naively adapting existing sample-wise attacks to this universal setting faces three core difficulties: (i) target supervision becomes high-variance due to \emph{triple randomization}, (ii) token-wise matching is unreliable because universality suppresses image-specific cues that would otherwise anchor alignment, and (iii) few-source per-target adaptation is highly initialization-sensitive, which can degrade the attainable performance. In this work, we propose TarVRoM-Attack, which stabilizes supervision via Target-View Aggregation with an Attention-Focused View, improves token-level reliability through alignability-gated Token Routing, and meta-learns a cross-target perturbation prior that yields stronger per-target solutions. Across commercial MLLMs, we boost unseen-image attack success rate by +23.7\% on GPT-4o and +19.9\% on Gemini-2.0 over the strongest universal baseline.
\end{abstract}


\section{Introduction}

\begin{figure*}[t]
  \begin{center}
    \includegraphics[width=\linewidth]{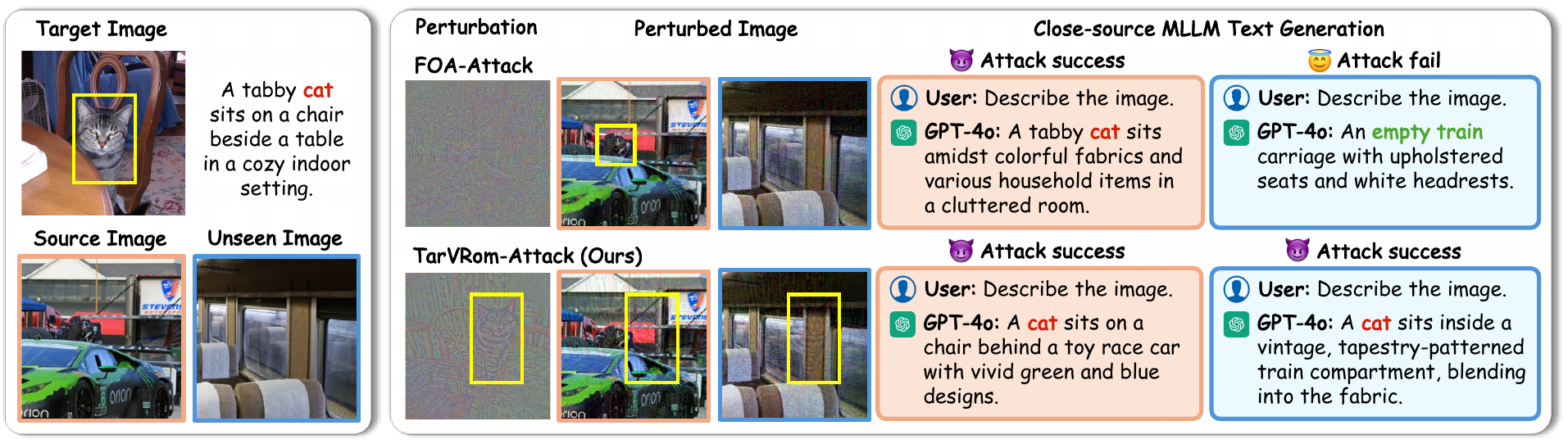}
    \vspace{-8mm}
    \caption{
       Comparison of targeted adversarial examples generated by FOA-Attack~\cite{jia2025adversarial} and our TarVRoM-Attack on the source image (captions in \textcolor{orange}{orange box}) and an unseen arbitrary image (captions in \textcolor{blue}{blue box}).
       Both methods succeed on the source images. However, FOA-Attack relies on local shadow cues and fails to transfer, while our TarVRoM-Attack consistently induces the target concept (``cat'') across heterogeneous backgrounds. Best viewed via zoom-in.
    }
    \label{fig:intro}
  \end{center}
  \vspace{-4mm}
\end{figure*}

Building upon the significant advancements in Large Language Models (LLMs) \cite{llama,qwen}, Multi-modal Large Language Models (MLLMs) have recently attracted substantial attention \cite{minigpt4,deepseek-vl,li2025otter}. 
However, despite their potential, the security of MLLMs remains a critical challenge, as existing models have been shown to be susceptible to adversarial attacks \cite{jiang2025survey}. In particular, targeted adversarial attacks are of special concern, as they aim to deliberately induce specific, attacker-chosen incorrect outputs, in contrast to untargeted attacks that cause arbitrary prediction failures \cite{zhao2023evaluating,guo2024efficient}. 

Commercial closed-source MLLMs, \textit{e.g.,} GPT-4o \cite{gpt4o}, Claude-4.5 \cite{claude}, and Gemini-2.0 \cite{gemini}, are not immune to such vulnerabilities.
In practice, attackers can exploit the transferability of adversarial perturbations crafted on accessible surrogate models to mount black-box attacks against proprietary systems \cite{lu2025pretrain,li2025a}.
Furthermore, perturbations optimized on open-source CLIP models have been shown to induce targeted mispredictions on closed-source MLLMs.
Nevertheless, the efficacy of these transfer-based targeted adversarial attacks is often hampered by a limited generalization to unseen images. 
Prior approaches employ a sample-wise optimization strategy \cite{jia2025adversarial, li2025a}, in which instance-specific perturbations are independently tailored to each input. Although effective in source images, such perturbations tend to overfit local visual patterns or specific semantic cues, failing to retain adversarial potency when applied to different images, as shown in Fig.~\ref{fig:intro}. This undermines the practical utility of targeted adversarial attacks, as optimizing unique perturbations for each input is computationally prohibitive or impractical in real-world settings. 
Consequently, we are motivated to explore an effective \textbf{Universal Targeted Transferable Adversarial Attack (UniTTAA) on closed-source MLLMs} with strong generalization across diverse visual inputs.

Naively extending sample-wise attacks to the universal setting is brittle for two reasons. First, prior methods~\cite{jia2025adversarial, li2025a} already rely on stochastic view sampling, where each update draws a source view and a target view independently through randomized cropping. In the universal setting, this is further compounded by stochastic source selection from a source pool. We refer to these three coupled randomness sources as \textit{triple randomization}, which inflates gradient variance and destabilizes optimization. Second, existing methods enforce token alignment indiscriminately, without specifying which source tokens should be matched to the target. This causes incidental textures and weakly related patches to be aligned as well, yielding spurious correspondences and noisy supervision. As a result, gradients may be driven by local view-specific patterns rather than stable target semantics.

In light of this, we propose \textbf{TarVRoM-Attack}, which aims to craft a universal perturbation capable of misleading closed-source MLLMs toward desired target outputs, independently of individual image characteristics.
To improve optimization stability, we introduce \textbf{Target-View Aggregation (TVA)} together with an \textbf{Attention-Focused View (AFV)}. Unlike prior methods that rely on a single potentially noisy local view~\cite{jia2025adversarial, li2025a}, TVA aggregates supervisory signals over multiple target views, thereby yielding a more reliable characterization of target-specific semantics. Moreover, we leverage a \textbf{Token Routing (TR)} mechanism to explicitly guide the model toward \textit{where} to attend during universal perturbation learning. By selectively emphasizing these alignable tokens, it is able to deliver more stable learning signals. 
Furthermore, we adopt a \textbf{Meta-Initialization (MI)} scheme that learns a target-agnostic perturbation prior by exposing the optimization process to a wide range of target concepts during meta-training. This initialization captures shared and transferable structures across targets, enabling the learned initial perturbation to generalize effectively to each target. Notably, with meta-initialization, our method achieves attack performance that is comparable to or even surpasses optimization from scratch with substantially more iterations (\textit{e.g.}, 50 steps w/ MI VS. 300 steps w/o MI).

The contributions of this work are threefold. \textit{First}, we present the first systematic study of UniTTAA against closed-source MLLMs, a substantially more challenging setting than previous sample-wise attacks.
\textit{Second}, we propose TarVRoM-Attack, a universal targeted transferable adversarial attack that jointly stabilizes universal learning via TVA with AFV, enhances informative supervision through TR with selective alignment, and learns a target-agnostic perturbation prior through MI. \textit{Third}, extensive experiments demonstrate that adversarial perturbations generated by our TarVRoM-Attack exhibit strong generalization to previously unseen images across multiple commercial MLLMs, while remaining competitive with sample-wise methods on seen source images. In particular, our approach improves upon the best baseline by an \textbf{absolute 23.7\%} on GPT-4o and \textbf{19.9\%} on Gemini-2.0 in attack success rate for unseen images.

\section{Related Work}

\noindent \textbf{MLLMs.}
Recent progress in LLMs \cite{llama,qwen} has spurred a growing interest in MLLMs, which integrate language understanding with visual modality \cite{qwenvl,minigpt4,li2025otter}. 
Representative models such as LLaVA \cite{llava} and DeepSeek-VL \cite{deepseek-vl} demonstrate impressive capabilities across a wide range of multimodal tasks, including image captioning \cite{image-captioning}, visual question answering \cite{vqa}, and visual complex reasoning \cite{reasoning}. 
In addition to open-source progress, several closed-source commercial MLLMs, such as GPT-4o \cite{gpt4o}, Claude-4.5 \cite{claude}, and Gemini-2.0 \cite{gemini}, have also been widely adopted. 

\noindent\textbf{Transferable adversarial attacks} craft adversarial examples on accessible surrogate models and transfer them to unseen victims. Prior work improves transferability mainly via: (i) optimization refinements for more model-agnostic gradients (FGSM/iterative variants, momentum, smoothing)~\cite{kurakin2018adversarial,dong2018boosting,zhu2023boosting}; (ii) input diversification with stochastic transforms (resize-pad, multi-scale, translation, mixing, block-wise)~\cite{xie2019improving,dong2019evading,chen2024diffusion,wang2021admix}; and (iii) feature-level manipulation (highlighting influential neurons or aligning intermediate patterns)~\cite{fda,fia,naa,wei2023adaptive,wang2024boosting,liang2025improving}.





\noindent\textbf{Targeted Attack on MLLMs.}
Attacks on MLLMs can be untargeted (degrading responses) or targeted (steering outputs to a specified goal). Recent work increasingly studies \emph{transferable targeted} attacks, including surrogate-based transfer~\cite{zhao2023evaluating}, diffusion-guided optimization~\cite{guo2024efficient}, and frequency/ensemble-enhanced black-box methods~\cite{dong2023robust}. Transfer is further boosted by stronger surrogates/generators~\cite{zhang2025anyattack}, lightweight stochastic augmentations~\cite{li2025a}, and joint global--local feature alignment for proprietary models~\cite{jia2025adversarial,hu2025transferable}.

\noindent\textbf{Our Motivation.} 
While universal transferable attacks on MLLMs have attracted increasing attention in recent years \cite{chen2020universal,zhang2024universal,zhou2025vanish}, we focus on a substantially more challenging \textit{targeted universal} setting, in which a single perturbation must consistently steer model output toward a specified target. Unlike X-Transfer \cite{huang2025x}, whose targets are limited to 10 fixed text descriptions, our framework supports arbitrary target images, providing a more realistic target space for vision-centric tasks.

\section{Preliminary and Problem Formulation}
\label{sec:setting}
Existing transfer-based targeted adversarial attacks against closed-source MLLMs operate in a \textit{sample-wise} manner~\cite{jia2025adversarial, li2025a}: they optimize an instance-specific perturbation for each input image $\bm{x}$.
While effective for the given sample, the perturbation $\bm{\delta}$ typically fails to generalize to unseen images $\bm{x}' \neq \bm{x}$, requiring computationally expensive re-optimization for every new input $\bm{x}'$. We instead study a \textbf{Universal Targeted Transferable Adversarial Attack (UniTTAA)}, which learns a \emph{single} input-agnostic perturbation that, for \emph{arbitrary} inputs, steers the victim MLLM’s output (or visual embedding) toward the target image $\bm{x}_{\mathrm{tar}}$. Our objective is as follows.

\begin{definition}[\textbf{UniTTAA}]
\label{def:universal_attack_rewrite}
Given a target image $\bm{x}_{\mathrm{tar}}\in\mathbb{R}^{H\times W\times C}$ and a feasible space $\mathcal{S}$ (\textit{e.g.,} $\|\bm{\delta}\|_{\infty}\le \tfrac{16}{255}$),
a {universal targeted transferable} perturbation is a {single} perturbation $\bm{\delta}$ (shared across inputs) that, when added to an arbitrary clean image, drives {unknown victim models} to match the target in the specified representation space:
\begin{equation}
\label{eq:def_attack_main}
\bm{\delta}^{\star}\in
\arg\min_{\underline{\bm{\delta}\in\mathcal{S}}}\;
\mathbb{E}_{\underline{\bm{x}\sim\mathcal{D}}}\;
\mathbb{E}_{\underline{\hat f\sim\hat{\mathcal{F}}}}
\Big[
\mathcal{L}_{\mathrm{eval}}\!\big(\bm{x}+\bm{\delta},\,\bm{x}_{\mathrm{tar}};\hat f\big)
\Big],
\end{equation}
where $\bm{x}$ is an arbitrary clean image sampled from a natural distribution $\mathcal{D}$, $\hat{\mathcal{F}}$ is a family of (unknown) victim MLLMs, and
$\mathcal{L}_{\mathrm{eval}}(\cdot)$ is an external, output-level discrepancy that compares the model response to the perturbed image $\hat f(\bm{x}+\bm{\delta})$ with that to the target image $\hat f(\bm{x}_{\mathrm{tar}})$,
\textit{e.g.,} via a GPT-based judge for caption similarity or keyword matching.
The underlined $\bm{\delta}$ highlights \textbf{universal} (one perturbation per target), while the underlined expectations over $\bm{x}$ and $\hat f$ highlight \textbf{transfer} (across arbitrary images and models).
\end{definition}


Since victim MLLMs are typically \textbf{closed-source}, we follow \cite{jia2025adversarial, li2025a} and optimize on an ensemble of image encoders
$\mathcal{F}=\{f_{\theta_1},\ldots,f_{\theta_t}\}$ from vision-language pretrained models to obtain transferable features.
Given a small source image set $\mathcal{X}=\{\bm{x}_j\}_{j=1}^{n}$, we learn \textbf{one} universal perturbation per target $\bm{x}_{\mathrm{tar}}$ by solving the following empirical objective:
\begin{proposition}[\textbf{Empirical Optimization for Universal Targeted Transfer}]
\label{prop:empirical_opt_short}
Given a small source image set $\mathcal{X}=\{\bm{x}_j\}_{j=1}^{n}$ and the encoder ensemble $\mathcal{F}=\{f_{\theta_i}\}_{i=1}^{t}$, we solve
\begin{equation}
\label{eq:emp_obj_short}
\bm{\delta}_s \in \arg\min_{\bm{\delta}\in\mathcal{S}}\;
\sum_{j=1}^{n}\sum_{i=1}^{t}
\mathcal{L}_{\mathrm{train}}\!\big(\bm{x}_{j}+\bm{\delta},\,\bm{x}_{\mathrm{tar}}; f_{\theta_i}\big),
\end{equation}
where $f_{\theta_i}$ extracts image features and $\mathcal{L}_{\mathrm{train}}(\cdot)$ is a feature-space surrogate that aligns $\bm{x}_j+\bm{\delta}$ with $\bm{x}_{\mathrm{tar}}$ under $f_{\theta_i}$, \textit{e.g.,} cosine/MSE distance.
Eq.~\eqref{eq:emp_obj_short} is a tractable surrogate of Def.~\ref{def:universal_attack_rewrite}, and we evaluate transfer to unseen images and unknown victim MLLMs $\hat f\sim\hat{\mathcal{F}}$ using $\mathcal{L}_{\mathrm{eval}}(\bm{x}+\bm{\delta}_s,\,\bm{x}_{\mathrm{tar}};\hat f)$.
\end{proposition}

\section{Methodology}
\label{sec:methods}
\textbf{Overview.}
Under the UniTTAA formulation in Sec.~\ref{sec:setting}, we aim to learn a \emph{single} perturbation $\bm{\delta}$ for a given target image $\bm{x}_{\mathrm{tar}}$ using surrogate models, and evaluate its transfer to \emph{unseen} source images and \emph{unknown} victim MLLMs. Our TarVRoM-Attack is detailed in the following subsections.
Sec.~\ref{sec:MCA} proposes TVA to reduce optimization variance by aggregating supervisory signals across a multi-view target set. Sec.~\ref{sec:TR} introduces TR to enhance informative supervision by routing alignable tokens while regularizing non-alignable ones. Sec.~\ref{sec:meta} learns a meta-initialization that enables scalable target adaptation. Finally, Sec.~\ref{sec:twostage} summarizes the overall two-stage optimization of our TarVRoM-Attack.



\subsection{Stabilizing Supervision via Target-View Aggregation (TVA)}
\label{sec:MCA}

\begin{figure}
    \centering
    \includegraphics[width=\linewidth]{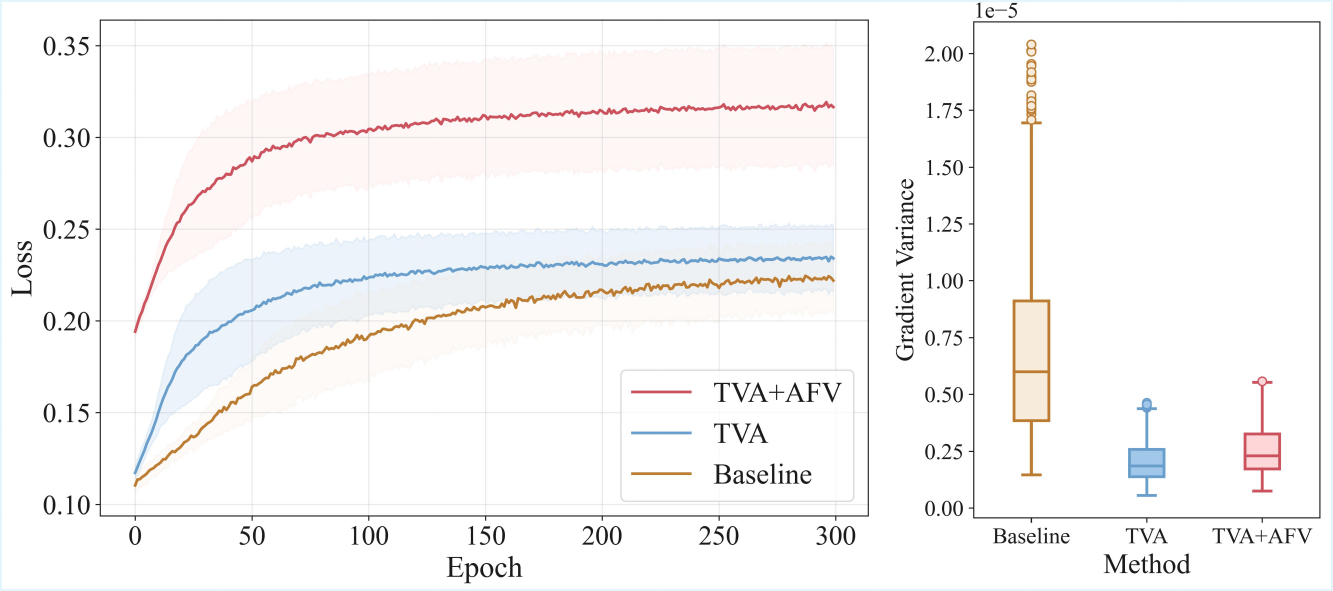}
    \vspace{-8mm}
    \caption{
    (Left) Comparison of mean loss curves with variance shading over 300 epochs, where the TVA and TVA+AFV variants exhibit improved convergence behavior relative to the baseline. (Right) Illustration of gradient variation, indicating that the proposed methods effectively reduce gradient stochasticity.}
    \label{fig:loss}
    \vspace{-4mm}
\end{figure}

\noindent \textbf{Why sample-wise approaches become unstable in the universal Setting.}
Existing targeted transferable attacks are predominantly \textit{sample-wise}, performing token-level alignment between a randomly sampled source view and a \textit{single} randomly sampled target view at its original resolution \cite{jia2025adversarial,li2025a}.
When extended to the UniTTAA setting, this update strategy becomes significantly more challenging to optimize. For a given target image, a pool of source images is available, and at each optimization step, a \textit{triple-randomization} occurs: a source image is randomly sampled from the pool, and both the source and target images are independently transformed into stochastic local views. As a result, the gradient at each update is dominated by view-specific local token fluctuations, leading to a high-variance estimate of the desired objective. Consequently, the update may degenerate into a stochastic walk rather than progressing toward the expected optimum, leading to slow convergence and elevated gradient variance, as illustrated in yellow in Fig.~\ref{fig:loss}.

To stabilize universal optimization, we propose \textbf{Target-View Aggregation (TVA)}, which replaces the conventional one-view-per-step target supervision with an aggregated estimate over a multi-view target set. By integrating supervisory signals across multiple target views, TVA promotes a more robust characterization of target-specific semantics and reduces over-reliance on any single potentially noisy view. Prop.~\ref{prop:mc_unbiased_MCA} further provides theoretical support for TVA by formalizing it as an unbiased Monte Carlo estimator with reduced variance. Moreover, we introduce an \textbf{Attention-Focused View (AFV)} as a persistent anchor, since regions with high attention score contain richer semantic information, providing more informative signals that facilitate convergence to the optimum.

\begin{proposition}[\textbf{Monte Carlo Unbiasedness and Variance Reduction}]
\label{prop:mc_unbiased_MCA}
Let $v \sim p(v)$ denote a 
randomly sampled view from the target image $\bm{x}_{\mathrm{tar}}$ at its original resolution, and define the per-view objective as:
\begin{equation}
\ell(\bm\delta;v):= \mathcal{L}\!\left(\bm\delta, v(\bm{x}_{\mathrm{tar}})\right),~
J(\bm\delta):= \mathbb{E}_{v\sim p(v)}\big[\ell(\bm\delta;v)\big].
\end{equation}
Given i.i.d.\ views $\{v_i\}_{i=1}^m$ from $p(v)$, we consider the multi-view estimator given below:
\begin{equation}
\begin{aligned}
\widehat{J}_m(\bm\delta):=\frac{1}{m}\sum_{i=1}^m \ell(\bm\delta;v_i),
~~\widehat{\bm g}_m(\bm\delta):= \nabla_{\bm\delta}\widehat{J}_m(\bm\delta)
        .
\end{aligned}
\end{equation}
Assume $\nabla_{\bm\delta}\ell(\bm\delta;v)$ is integrable and differentiation can be
interchanged with expectation. Then, we have:
\begin{equation}
\begin{aligned}
\mathbb{E}\big[\widehat{J}_m(\bm\delta)\big] &= J(\bm\delta),
~~\mathbb{E}\big[\widehat{\bm g}_m(\bm\delta)\big] = \nabla_{\bm\delta} J(\bm\delta),\\
\mathrm{Var}\big(\widehat{\bm g}_m(\bm\delta)\big) &= \frac{1}{m}\,
\mathrm{Var}\big(\nabla_{\bm\delta}\ell(\bm\delta;v)\big).
\end{aligned}
\end{equation}
\end{proposition}

\begin{remark}
\label{rem:mc_MCA}
 Prop.~\ref{prop:mc_unbiased_MCA} formalizes that replacing ``one-view-per-step'' with a multi-view target set yields an {unbiased} estimate of the distribution-level objective,
while reducing gradient variance by a factor of $1/m$. This insight motivates TVA as a principled strategy to stabilize universal targeted optimization under limited steps.
\end{remark}
\noindent\textbf{Implementation.} We instantiate TVA by representing the target with a small set of views at the original resolution.
Specifically, we first sample $m-1$ views
$\mathcal{V}(\bm{x}_\mathrm{tar})=\{v_i(\bm{x}_\mathrm{tar})\}_{i=1}^{m-1}$ at random.
To provide a stable anchor signal, we add an attention-focused view
$v_{\mathrm{attn}}(\bm{x}_\mathrm{tar})$, 
obtained by anchoring a square window at the peak of the surrogate model’s final-layer attention map, setting its side length to the nearest image boundary (with a minimum size), clipping it within the image, and resizing it.
We denote the target-view set by:
\begin{equation}
\label{eq:establish-set}
\mathcal{V}^+(\bm{x}_\mathrm{tar})
\triangleq
\mathcal{V}(\bm{x}_\mathrm{tar})\cup\{v_{\rm attn}(\bm{x}_\mathrm{tar})\}.
\end{equation}
For each target view $v_{\mathrm{tar}}\in\mathcal{V}^+(\bm{x}_\mathrm{tar})$, 
we compute global features for the $i$-th surrogate encoder $f_{\theta_i}$:
\begin{equation}
\mathbf{g}_{\mathrm{tar}}^{\,i}=f_{\theta_i}^{g}(v_{\mathrm{tar}}), \qquad
\mathbf{g}_{\mathrm{adv}}^{\,i}=f_{\theta_i}^{g}(v_{\mathrm{adv}}),
\end{equation}
where $f_{\theta_i}^{g}(\cdot)$ are the global-level outputs and $v_{\mathrm{adv}}=v(\bm{x}+\bm{\delta})$ is an adversarial source view. We then define the global cosine alignment term as follows.
\begin{equation}
\mathcal{L}_{\mathrm{global}}^{i}
=\cos\!\big(\mathbf{g}_{\mathrm{adv}}^{\,i}, \mathbf{g}_{\mathrm{tar}}^{\,i}\big).
\end{equation}
In this way, the global alignment term is no longer tied to a single potentially noisy view,
but instead matches an aggregated target estimate whose unbiasedness and variance-reduction
properties are justified by Prop.~\ref{prop:mc_unbiased_MCA}, making TVA a principled and
stable supervision mechanism.

\subsection{Token Routing (TR) via Alignability Gating}
\label{sec:TR}
 
In the universal targeted setting, enforcing token-wise alignment equally across all source patches is unreliable, since a single perturbation must generalize across diverse images and thus cannot preserve instance-specific cues. As a result, indiscriminate alignment at the token level tends to create spurious correspondences and noisy supervision. 

\noindent\textbf{Exploit what aligns.} To this end, we propose \textbf{Token Routing (TR)}, which recognizes that not all source tokens are equally informative for learning a universal perturbation. For a given adversarial source view, only a subset of tokens exhibits structural or semantic compatibility with the target. By emphasizing these alignable tokens, TR provides stable and transferable supervision by explicitly guiding the model to focus on the most relevant features.

\noindent\textbf{Implementation.}
For the $i$-th surrogate encoder, let
\begin{equation}
\begin{aligned}
\mathbf{P}^{\,i}_{\mathrm{tar}}=\{\mathbf{p}_k^{\,i}\}_{k=1}^{K}
&=\mathrm{KMeans}\!\left(f_{\theta_i}^{t}(v_{\mathrm{tar}})\right),\\
\mathbf{T}_{\mathrm{adv}}^{\,i}=\{\mathbf{t}_n^{\,i}\}_{n=1}^{N}
&=\mathrm{KMeans}\!\left(f_{\theta_i}^{t}(v_{\mathrm{adv}})\right),
\end{aligned}
\end{equation}
denote the token features of the adversarial source view and 
the target token prototypes obtained by applying K-means to the TVA target features, where $f_{\theta_i}^{t}(\cdot)$ are the token-level outputs, $k$ denotes the target prototype index, and $n$ denotes the source token index. We first measure the alignability of each adversarial token by its maximum cosine similarity to the target prototypes:
\begin{equation}
r_n^{i}=\max_{k}\cos\!\big(\mathbf{t}_n^{\,i},\mathbf{p}_k^{\,i}\big).
\end{equation}
The score is then converted into a soft routing weight
\begin{equation}
w_n^{i}
=\sigma\!\left(\frac{r_n^{i}-\gamma}{\beta}\right),
\end{equation}
where $\sigma(\cdot)$ is sigmoid function, $\gamma$ sets its threshold and $\beta$ controls the sharpness of the gate. In this way, tokens that are better aligned with the target receive larger transport mass, whereas less relevant tokens are naturally down-weighted.

Based on these routing weights, we formulate local alignment as a weighted optimal transport problem. Specifically, we define the cosine similarity matrix between target prototypes and adversarial tokens as
\begin{equation}
\!\!S^{i}_{kn}\!=\!\cos\!\big(\mathbf{p}_k^{\,i},\mathbf{t}_n^{\,i}\big),
\,
C^{i}_{kn}\!=\!1\!-\!S^{i}_{kn},\,
K^{i}_{kn}\!=\!\exp\!\left(\!-\frac{C^{i}_{kn}}{\theta}\!\right),
\end{equation}
where $C^{i}_{kn}$ is the transport cost and $K^{i}_{kn}$ is a kernel with $\theta=0.1$. The source marginal $\mathbf{x}^{i}$ over adversarial tokens is set uniformly, while the target marginal $\mathbf{y}^{i}$ over prototypes is induced by the routing weights,
\begin{equation}
\mathbf{x}^{i}\!=\!\left(\frac{1}{K},\ldots,\frac{1}{K}\right),~\mathbf{y}^{i}\!=\!\frac{\mathbf{w}^{i}}{\sum_{n=1}^{N} w_n^{i}},~\mathbf{w}^{i}\!=\!(w_1^i,\dots,w_N^i).
\end{equation}
Then, based on the kernel, we obtain the transport plan $\Pi^i_{kn}=\mathrm{Sinkhorn}( K^i_{kn},\mathbf x^i,\mathbf y^i)$ via Sinkhorn normalization\cite{cuturi2013sinkhorn}, and define the routed local alignment objective as
\begin{equation}
\mathcal{L}_{\mathrm{route}}^{i}
=\sum_{k=1}^{K}\sum_{n=1}^{N}\Pi^{i}_{kn} S^{i}_{kn}.
\end{equation}
Finally, for an ensemble of surrogate encoders, we combine the global and routing terms under adaptive model weighting and optimize
\begin{equation} \label{eq:overall_loss}
\mathcal{L}
=\sum_{i=1}^{M} W_i\!\left(
\mathcal{L}_{\mathrm{global}}^{i}
+\mathcal{L}_{\mathrm{route}}^{i}
\right),
\end{equation} 
where $W_i$ denotes the adaptive weight of the $i$-th surrogate~\cite{jia2025adversarial}. Overall, TVA provides stable target-level supervision, while TR further refines this supervision at the token level by routing transport mass toward more alignable local structures, yielding a more robust and target-consistent optimization objective.

\subsection{Meta-Initialization for Scalable Target Adaptation}
\label{sec:meta}
UniTTAA is inherently a \emph{many-target} problem: as new targets arrive, the attacker must produce a \emph{per-target} universal perturbation.
In practice, each target typically comes with only a small support set of source images for adaptation~\cite{mopuri2018generalizable,zhang2021data,Zhang_2025_CVPR}, \textit{i.e.}, $N\le 20$.
This \emph{few-source} regime makes per-target optimization highly sensitive to initialization: with limited supervision, training from a zero initialization $\bm{\delta}_0$ is prone to drifting toward target- or sample-specific shortcuts, which weakens transfer and degrades the attainable performance.
Prior work has also shown that, under few-shot adaptation settings, a well-learned initialization can act as a transferable prior and substantially improve the final solution quality~\cite{xia2024transferable, zhou2025darkhash, yin2023generalizable}.
Motivated by these works, we do not treat each target in isolation.
Instead, we learn a \emph{generalizable prior} from \emph{many meta tasks}, so that the initialization already encodes update directions that consistently support targeted universal transfer across diverse targets.
We therefore propose a \emph{target-as-task meta-initialization} that provides a transferable warm start and yields stronger per-target performance in practice. 
Specifically, we sample $N$ sources for each target $\tau_b$ from $\mathcal D_s=\{\bm x_{i}\}_{i=1}^{|\mathcal D_s|}$ as below:
\begin{equation}
\label{eq:sampleN}
\begin{aligned}
    \mathcal I_{\tau_b}&\subseteq \{1,\dots,|\mathcal D_s|\},\;
|\mathcal I_{\tau_b}|=N,\\
\mathcal I_{\tau_b}\!\sim\! \mathrm{Uniform}&\left(\!\Big \{ \mathcal I\subseteq \{1,\dots ,|\mathcal D_s|\}:\ |\mathcal I|=N\Big\}\!\right), \\
\mathcal X_{\tau_b}\triangleq& \{\bm x_{i}:\ i\in \mathcal I_{\tau_b}\}\in \mathbb R^{N\times H\times W\times C}.
\end{aligned}
\end{equation}

\noindent\textbf{Target-as-Task Meta-initialization (MI).}
We meta-learn an initialization $\bm\delta_0$ that is optimized for adaptation. After a small number of inner steps on a few-shot support set, it should yield a strong per-target perturbation.
We treat each target $\bm x_{\rm tar}\sim\mathcal D_{\rm tar}$ as a task $\tau$ and learn $\bm\delta_0$ via a first-order Reptile update \cite{nichol2018first}.
Crucially, the inner adaptation operator is \emph{aligned with} our test-time multi-view targeting procedure, employing the same view construction and the same alignment loss, so meta-learning directly optimizes the post-adaptation objective, making $\bm\delta_0$ an \emph{update prior} that amortizes per-target optimization.

\noindent\textbf{Meta Objective.} For task $\tau$, the multi-view target optimization applies $I$ steps of a constrained attacker update on a small support set $\mathcal X_{\tau}=\{\bm{x}_1, \cdots, \bm{x}_n\}$. We denote this inner adaptation operator by \begin{equation} \bm\delta_\tau \;=\; U^{(I)}_\tau(\bm\delta_0), \label{eq:inner_operator} \end{equation} where $\bm\delta_0$ is the shared zero-initialization and $U^{(I)}_\tau(\cdot)$ includes the stochastic sampling and target-view construction. The meta goal is to learn $\bm\delta_0$ that yields strong \emph{post-adaptation} performance under the per-target budget: \begin{equation} J(\bm\delta_0) =\mathbb{E}_{\tau\sim p(\mathcal{D}_\mathrm{tar})} \Big[ \mathcal{L}_\tau\big(U^{(I)}_\tau(\bm\delta_0)\big) \Big], \label{eq:meta_obj} \end{equation} where $\mathcal{L}_\tau(\cdot)$ is defined to be consistent with $\mathcal{L}$ in Eq.~\eqref{eq:overall_loss}.

\noindent\textbf{Reptile Meta-update.}
To optimize the expected post-adaptation objective in Eq.~\eqref{eq:meta_obj} without backpropagating through the $I$ inner steps, we adopt the first-order Reptile update~\cite{nichol2018first}.
At each meta iteration, we sample a mini-batch of tasks $\{\tau_b\}_{b=1}^{B}$, run the inner adaptation in Eq.~\eqref{eq:inner_operator} to obtain the post-adapt perturbations $\{\bm\delta_{\tau_b}\}$, and move the initialization toward their average:
\begin{equation}
\bm\delta_0 \leftarrow
\Pi_{\|\bm\delta\|_\infty\le \epsilon}\!\left(
\bm\delta_0 + \eta(\bar{\bm\delta}-\bm\delta_0)
\right),~
\bar{\bm\delta}=\tfrac{1}{B}\sum_{b=1}^B \bm\delta_{\tau_b}.
\label{eq:reptile_update}
\end{equation}
{$\Pi_{\|\bm\delta\|_\infty\le \epsilon}(\cdot)$ denotes the projection (clamp) in terms of $\ell_{\infty}$-norm with budget $\epsilon$.}
Intuitively, $(\bar{\bm\delta}-\bm\delta_0)$ aggregates task-specific adaptation directions, yielding a first-order surrogate for the post-adaptation risk in Eq.~\eqref{eq:meta_obj}.
Corresponding Proposition and Remark 
clarify how the Reptile update in Eq.~\eqref{eq:reptile_update} serves as a first-order procedure for optimizing the expected post-adaptation objective in Eq.~\eqref{eq:meta_obj}.


\begin{algorithm}[t]
\caption{\textsc{InnerUpdate} for TarVRoM-Attack}
\label{alg:innerupdate}
\footnotesize
\begin{algorithmic}[0] 
\FUNCTION{InnerUpdate($\bm{\delta}, \tau_b, \mathcal{X}, \alpha$)}
\REQUIRE clean sources $\mathcal{X}$, target image $\tau_b$
\REQUIRE init perturbation $\bm{\delta}$, step size $\alpha$
\STATE $\mathcal{V}^+\leftarrow
\mathcal{V}(\tau_b)\cup\{v_{\rm attn}(\tau_b)\}$ via Eq. \eqref{eq:establish-set}
\FORALL{$(\bm{x}, v)\in \mathcal{X}\times\mathcal{V}^{+}$}
\STATE $\mathcal{L}\leftarrow \mathcal{L}(\bm{x},\bm{\delta},v)$ via Eq.~\eqref{eq:overall_loss}
\STATE $\bm{\delta}\leftarrow \bm{\delta}$ updated with $\mathcal{L},\alpha$ via Eq.~\eqref{eq:innerupdate}
\ENDFOR
\STATE \textbf{return} $\bm{\delta}$
\ENDFUNCTION
\end{algorithmic}
\end{algorithm}

\subsection{Meta-to-Target Adaptation for TarVRoM-Attack}
\label{sec:twostage}
As shown in Alg.~\ref{alg:innerupdate}, for each inner update, we update $\bm\delta$ with a projected FGSM \cite{goodfellow2014explaining} based on the loss defined in Eq.~\eqref{eq:overall_loss}:
\begin{equation}
\label{eq:innerupdate}
    \bm{\delta}\gets \Pi_{\|\bm\delta\|_\infty\le \epsilon}\Big(\bm{\delta}+\alpha\cdot \mathrm{sign}\big(\nabla_{\bm{\delta}}\mathcal{L}\big)\Big).
\end{equation}
The overall optimization is summarized in Alg.~\ref{alg:meta_fast_adapt_uap}, where:
\noindent\textbf{Stage-1: Meta (Reptile) Training of $\bm{\delta}_0$.}
At meta epoch $e$, we sample sub-target tasks $\mathcal{D}_{\mathrm{tar}}^{\mathrm{sub}}=\{\tau_b\}_{b=1}^{B}\subset \mathcal{D}_{\mathrm{tar}}$.
For each task $\tau_b$, we construct a target-view set $\mathcal{V}^+(\tau_b)$.
Let $\bm{\delta}_{\tau_b}$ denote the inner-updated perturbation for $\tau_b$, and we then update the initialization $\bm\delta_0$ via Eq.~\eqref{eq:reptile_update}.

\noindent\textbf{Stage-2: Meta-to-target Adaptation to Each Target.}
For each target image $\tau_b\in\mathcal{D}_{\mathrm{tar}}$,
we build $\mathcal{V}^+(\tau_b)$ (Eq.~\eqref{eq:establish-set}) and start from
the learned initialization $\bm{\delta}_0$.
Running $M$ inner steps with the same projected sign update yields the per-target universal
perturbation $\bm{\delta}_{\tau_b}$, aiming to generalize across unseen samples and unknown closed-source MLLMs.

\begin{algorithm}[t]
\caption{Overall optimization of TarVRoM-Attack}
\label{alg:meta_fast_adapt_uap}
\footnotesize
\begin{algorithmic}[1]
\STATE \textbf{Input:} source pool $\mathcal{D}_s$, target pool $\mathcal{D}_{\mathrm{tar}}$, meta-init epochs $E$, meta-init task batch size $B$, meta-init inner steps $I$, meta-init step size $\eta$, adaptation steps $M$, adaptation step size $\alpha$.
\STATE \textbf{Output:} meta-init perturbations $\bm{\delta}_0$ and per-target perturbations 
$\{\delta_{\tau_b}\}_{\tau_b \in \mathcal D_{\mathrm{tar}}}$.
\STATE \textbf{Init:} $\bm{\delta}_0\gets \mathbf{0}$\\
\textcolor{teal}{\# Stage-1: Meta-initialization via Reptile Training of $\bm{\delta}_0$}
\FOR{$e=1$ \textbf{to} $E$}
    \STATE Sample meta-init tasks $
    \mathcal{D}_{\mathrm{tar}}^{\mathrm{sub}}=\
    \{\tau_b\}_{b=1}^{B}\subset \mathcal{D}_{\mathrm{tar}}$
    \FOR{$b=1$ \textbf{to} $B$}
    \STATE Sample $\mathcal{X}_{\tau_b}$ via~Eq.~\eqref{eq:sampleN}
    \FOR{$i=1$ \textbf{to} $I$}
        \STATE $\bm{\delta}_{\tau_b}\gets$ \textsc{InnerUpdate}($\bm{\delta}_0,\,\tau_b, \mathcal{X}_{\tau_b}, \eta$)
    \ENDFOR
    \ENDFOR
    \STATE Update meta initialization $\bm\delta_0$ via Eq.~\eqref{eq:reptile_update}
\ENDFOR

\textcolor{teal}{\# Stage-2: Meta-to-target Adaptation to Each Target}
\FORALL{$\tau_b\in\mathcal{D}_{\mathrm{tar}}$}
\STATE Sample $\hat{\mathcal{X}}_{\tau_b}$ via~Eq.~\eqref{eq:sampleN}
\FOR{$m=1$ \textbf{to} $M$}
    \STATE $\bm{\delta}_{\tau_b}\gets$ \textsc{InnerUpdate}($\bm{\delta}_0,\,\tau_b, \hat{\mathcal{X}}_{\tau_b}, \alpha$)
\ENDFOR
\ENDFOR
\STATE \textbf{Return} $\bm{\delta}_0$ and $\{\delta_{\tau_b}\}_{\tau_b \in \mathcal D_{\mathrm{tar}}}$
\end{algorithmic}
\end{algorithm}

\begin{table*}[t]\small
\centering
\captionsetup{font=small}
\caption{\textbf{Results on closed-source MLLMs}. Top: performance on unseen source samples not used to optimize the perturbation. Bottom: performance on seen source samples used during optimization.}
\label{main_tb}
\input{./tables/tbl1_universal_mllm}

\vspace{-4mm}
\end{table*}

\section{Experiments}
\noindent\textbf{Datasets.}
We follow \cite{dong2023robust,li2025a,jia2025adversarial} to select 100 target images from MSCOCO validation set~\cite{lin2014microsoft}. For each target, we sample 20 images for optimization and 30 disjoint unseen images for evaluation from NIPS 2017 Adversarial Attacks and Defenses Competition dataset~\cite{nips-2017-defense-against-adversarial-attack}, ensuring no data leakage.

\noindent\textbf{Competitive Methods.}
We include 5 baselines: AnyAttack~\cite{zhang2025anyattack}, M-Attack~\cite{li2025a}, and FOA-Attack~\cite{jia2025adversarial} represent sample-wise targeted transferable attack methods. We also include UAP~\cite{moosavi2017universal} as a fair reference via our reimplementation, where we replace the optimization used in FOA-Attack with the objective for targeted universal adversarial perturbations~\cite{hirano2020simple}. Finally, UnivIntruder~\cite{xu2025one} is a targeted universal attack that uses text prompts as targets.

\begin{figure}
    \centering
    \includegraphics[width=1.0\linewidth]{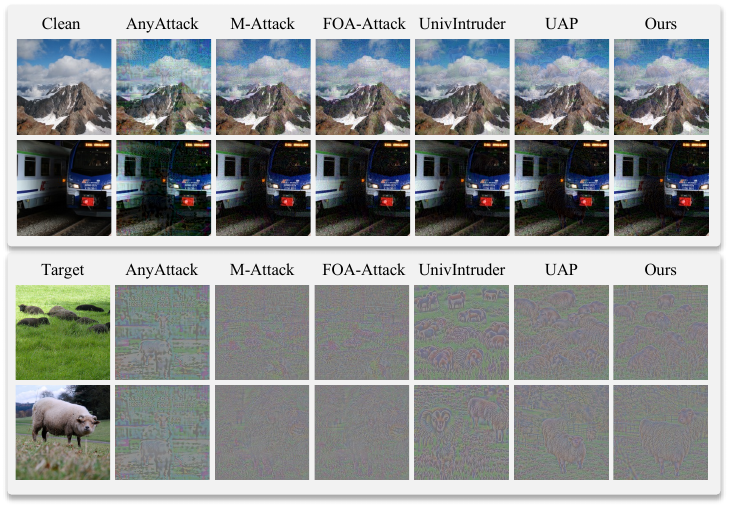}
\caption{Visualization of adversarial images/perturbations for unseen sample.}
    \label{fig:cases}
\end{figure}

\noindent\textbf{Evaluation Metrics and protocol.}
Following prior protocols \cite{li2025a}, we use an LLM-as-a-judge evaluation: the same closed-source model captions both target and adversarial images, and GPTScore measures their semantic similarity. We report attack success rate (ASR; similarity $>0.3$), average similarity (AvgSim), and keyword matching rates (KMR), where KMR$_a$/KMR$_b$/KMR$_c$ denote matching at least 1/2/3 of three annotated keywords. ASR under more thresholds is given in Appendix. Each table is split into two parts. The \textbf{top} part reports results on \emph{unseen} source samples, never used to optimize the universal perturbation for a target image. This is the \emph{primary} setting. For sample-wise methods, unseen-source results are obtained by optimizing on the seen source set for the same target and then directly transferring the fixed perturbation to held-out unseen sources, without further updates. The \textbf{bottom} part reports results on the \emph{seen} source samples used during optimization, showing fitting strength and the seen-unseen gap. Please see Appendix for \textbf{Implementation Details.}

\subsection{Comparisons results}

Tab.~\ref{main_tb} reports universal targeted transfer to GPT-4o, Gemini-2.0, and Claude under unseen and seen evaluations. In the \textbf{unseen} setting (top block), our method consistently yields the best universal transfer across models and metrics, demonstrating stronger universality. On GPT-4o, it improves ASR from 38.0\% (UAP) to 61.7\%, with higher KMR (\textit{e.g.}, KMR$_a$ 52.0\% vs.\ 37.5\%); Gemini-2.0 shows similar gains (56.7\% vs.\ 36.8\%, KMR$_a$ 52.6\% vs.\ 40.2\%), and the same trend holds on Claude. In the \textbf{seen} setting (bottom block), our method also achieves strong targeted steering with only one perturbation per target, clearly outperforming universal baselines and approaching sample-wise methods. For example, on GPT-4o it reaches 85.5\% ASR, versus 66.7\% for UAP and 15.0\% for UnivIntruder, narrowing the gap to FOA-Attack and outperforming AnyAttack; similar trends appear on Gemini-2.0. On Claude, it even surpasses all sample-wise baselines.\textbf{Additional results} on more closed-source MLLMs (\textit{e.g.,} GPT-5/5.2, Gemini-2.5/3, and Claude-Opus-4.5) are reported in the Appendix.

\subsection{Performance Analysis} \label{sec:perf}
\noindent\textbf{Few-source Trade-off with $N$.}
Tab.~\ref{tab:one-shot} varies the number of seen optimization samples $N$ for learning one target-specific universal perturbation. In the \textbf{unseen} setting, performance improves steadily with $N$ across all three closed-source models, and our method consistently outperforms the fair universal baseline UAP. In the \textbf{seen} setting, ASR peaks at small $N$ (usually $2$ or $5$), with slightly weaker fitting at larger $N$. This fit-generalize trade-off suggests that larger $N$ supplies more diverse source gradients, suppresses instance-specific shortcuts, and yields more source-invariant perturbations. Overall, our method remains effective with small $N$ while generalizing much better to unseen sources.

\begin{table}
    \centering
\captionsetup{font=small}
\caption{\textbf{Few-source optimization results for our TarVRoM-Attack}. We vary the number of seen optimization samples $N$ used to learn a single target-specific universal perturbation. }
\label{tab:one-shot}
\vspace{-2mm}
\scriptsize{
\resizebox{\linewidth}{!}{
\setlength\tabcolsep{5pt} 
\renewcommand\arraystretch{1.0} 
\begin{tabular}{cc||cc|cc|cc}
\hline\thickhline

\rowcolor{gray20}
 & &\multicolumn{2}{c|}{\textbf{GPT-4o}} & \multicolumn{2}{c|}{\textbf{Gemini}} & \multicolumn{2}{c}{\textbf{Claude}} \\
\cline{3-8}
\rowcolor{gray20}
\multirow{-2}{*}{$n$-source}&\multirow{-2}{*}{Method} & ASR & AvgSim & ASR & AvgSim & ASR & AvgSim \\
\hline\hline
\multicolumn{8}{l}{\textcolor{gray!90}{\textit{Performance on Unseen Test Samples}}} \\ 
& UAP&3.7&0.03 &4.0&0.03&3.0&0.03\\
\partialgray
\multirow{-2}{*}{$N=2$} &Ours & 9.0  & 0.05 & 9.0  & 0.04 & 2.3  & 0.02 \\ \hline
&UAP &9.0&0.04&6.7&0.04&5.3&0.03\\
\partialgray
\multirow{-2}{*}{$N=5$} &Ours & 27.0 & 0.13 & 25.0 & 0.11 & 7.7  & 0.03 \\ \hline
&UAP &23.3&0.10&16.7&0.10& 7.3&0.05\\
\partialgray
\multirow{-2}{*}{$N=10$}&Ours & 42.3 & 0.17 & 40.0 & 0.17 & 12.8 & 0.06 \\ \hline
&UAP &38.0&0.17&36.8&0.17&8.7&0.05\\
\partialgray
\multirow{-2}{*}{$N=20$} &Ours& \textbf{61.7} & \textbf{0.27} & \textbf{56.7} & \textbf{0.25} & \textbf{15.0} & \textbf{0.07} \\

\hline\hline

\multicolumn{8}{l}{\textcolor{gray!90}{\textit{Performance on Seen Samples (Used for Optimization)}}} \\
&UAP &90.0&0.55&75.0&0.45&10.0&0.05\\
\partialgray
\multirow{-2}{*}{$N=2$} &Ours & 95.0 & \textbf{0.63} & 85.0 & \textbf{0.53} & 15.0 & 0.06 \\ \hline
&UAP &72.0&0.41&62.0&0.33&18.0&0.11\\
\partialgray
\multirow{-2}{*}{$N=5$}&Ours  & \textbf{98.0} & 0.57 & \textbf{90.0} & 0.47 & \textbf{28.0} & \textbf{0.13} \\ \hline
&UAP&73.0&0.38&62.0&0.32&17.0&0.11 \\ 
\partialgray
\multirow{-2}{*}{$N=10$}&Ours & 87.0 & 0.47 & 89.0 & 0.44 & 23.0 & 0.11 \\ \hline
&UAP &66.7&0.32&61.3&0.29&13.8&0.07\\ 
\partialgray
\multirow{-2}{*}{$N=20$}&Ours & 85.5 & 0.39 & 75.5 & 0.36 & 25.0 & \textbf{0.13} \\

\hline\thickhline
\end{tabular}}}
\vspace{-4mm}
\end{table}

\noindent\textbf{Sample Visualization. }
Fig.~\ref{fig:cases} compares adversarial images and perturbations on unseen sources under the same bound. Our method most faithfully preserves natural appearance while consistently steering outputs toward the target, reflecting stronger cross-sample generalization. Its perturbation maps are also more structured and semantically transferable, in line with the quantitative results. More visualizations, analyses, and MLLM responses are given in Appendix. 

\begin{table}[t]
\caption{Ablation of Target-View Aggregation (TVA), Attention-Focused View (AFV), and Token Routing Universal Targeting (TR) in our attack. }
\renewcommand\arraystretch{1.0} 
    \label{tab:ab}
    \input{./tables/tbl3_ablation_mca}
\end{table}
\begin{table}[t]\small
\centering
\captionsetup{font=small}
\caption{\textbf{Effect of the number of target views $m$ in target-view aggregation.} We report the Attack Success Rate (ASR) and Average Similarity (AvgSim) with varying view numbers.}
\label{tab:MCA}
\renewcommand\arraystretch{1.0} 
\input{./tables/tbl4_crops_m}
\end{table}
\begin{table}[t]\small
\centering
\captionsetup{font=small}
\caption{\textbf{Comparison of results with and without meta-initialization.} The upper half uses our meta-initialized perturbation $\bm\delta_0$ as the starting point for Stage-2, while the lower half (w/o meta-init) starts from a zero initialization.}
\label{tab:fast}
\renewcommand\arraystretch{1.0} 
\input{./tables/tbl5_fast_adaptation}
\end{table}

\subsection{Ablation Study} \label{sec:abl}

Tab.~\ref{tab:ab} shows that TVA, AFV, and TR provide complementary gains on both seen and unseen splits. From the universal baseline, each component alone improves ASR/AvgSim, with larger benefits on harder models (notably Claude). The largest boost comes from combining TVA+AFV, indicating that diverse views plus a salient view yield stronger and more stable target supervision. Adding TR further improves transfer in most cases by filtering non-alignable token gradients and improving alignable learning, giving the best overall results. 


\noindent\textbf{Effect of Target-View Aggregation.}
We ablate TVA and its view number $m$ in Tab.~\ref{tab:MCA}. 
Overall, increasing $m$ strengthens targeted transfer on both seen and unseen samples, validating that TVA improves target-semantic alignment for universal optimization. 
We also observe diminishing returns (and occasional mild fluctuations) when $m$ becomes large, consistent with the variance reduction perspective in Theorem~\ref{prop:mc_unbiased_MCA}. More analysis is in Appendix.


\noindent\textbf{Ablation on Meta-Initialization.}
Tab.~\ref{tab:fast} shows that our meta-initialization markedly speeds up Stage-2 adaptation and improves generalization. Starting from the meta-initialized $\bm\delta_0$, only 50 epochs already achieves strong seen performance (\textit{e.g.,} ASR 77.5\% / 70.0\% / 24.0\% on GPT-4o / Gemini / Claude), approaching the best baseline while using far fewer updates. Under the same small budget, it delivers substantially higher unseen performance (ASR 54.7\% / 49.0\% / 11.8\%), clearly surpassing the w/o-meta counterpart and existing baselines. In contrast, removing meta-initialization yields a less favorable starting point for Stage-2 and consistently underperforms across all epoch budgets, especially in the low-epoch regime, leading to weaker cross-sample transfer. 

\noindent\textbf{More discussions.} Appendix analyzes \textbf{several defense methods} and shows that our attack remains effective under practical defensive preprocessing. Appendix studies the \textbf{impact of $\epsilon$} and demonstrates consistently strong performance across varying perturbation constraints. Finally, Appendix evaluates \textbf{unseen target adaptation}, where our method generalizes well to new targets that never appear during MI (Stage-1).

\section{Conclusion}
This work makes the first study of \emph{universal targeted transferable} adversarial attack on closed-source MLLMs. We propose \textsc{TarVRoM-Attack}, a two-stage method that learns a meta-initialized perturbation from a few source samples for scalable target adaptation and then performs meta-to-target adaptation to produce target-specific universal perturbations. To improve stability and universality, our method integrates Target-View Aggregation with Attention-Focused View 
together with alignability-gated token routing to focus updates on alignable structures. Extensive experiments show strong targeted steering and robust transfer on commercial MLLMs.

\bibliographystyle{IEEEtran}
\bibliography{example_paper_abb}

\end{document}

%% file: tables/tbl1_universal_mllm.tex
\vspace{-2mm}
\scriptsize{
\resizebox{\linewidth}{!}{
\setlength\tabcolsep{2pt} 
\renewcommand\arraystretch{1.0} 
\begin{tabular}{l||ccccc|ccccc|ccccc}
\hline\thickhline

\rowcolor{gray20}
 & \multicolumn{5}{c|}{\textbf{GPT-4o}} & \multicolumn{5}{c|}{\textbf{Gemini-2.0}} & \multicolumn{5}{c}{\textbf{Claude}} \\
\cline{2-16}

\rowcolor{gray20}
\multirow{-2}{*}{Method} 
 & $\text{KMR}_a$ & $\text{KMR}_b$ & $\text{KMR}_c$ & ASR  & AvgSim & $\text{KMR}_a$ & $\text{KMR}_b$ & $\text{KMR}_c$ & ASR  & AvgSim & $\text{KMR}_a$ & $\text{KMR}_b$ & $\text{KMR}_c$ & ASR  & AvgSim \\
\hline\hline

\multicolumn{16}{l}{\textcolor{gray!90}{\textit{Performance on Unseen Test Samples}}} \\

AnyAttack~\cite{zhang2025anyattack}& 8.0 & 3.0 & 0.1 & 6.1 & 0.04 & 8.9 & 3.8 & 0.2 & 6.4 & 0.04 & 5.6 & 2.6 & 0.2 & 5.3 & 0.03\\

\rowcolor{gray10}
M-Attack~\cite{li2025a}& 4.6 & 2.1 & 0.1 & 3.4 & 0.02 & 5.0 & 2.3 & 0.2 & 3.0 & 0.02 & 4.3 & 1.8 & 0.1 & 2.6 & 0.02 \\

FOA-Attack~\cite{jia2025adversarial} & 4.5 & 1.9 & 0.1 & 3.3 & 0.02 & 5.6 & 2.2 & 0.2 & 3.1 & 0.02 & 4.2 & 1.7 & 0.1 & 2.6 & 0.02 \\

\hdashline
\rowcolor{gray10}
UAP~\cite{moosavi2017universal} & 37.5 & 23.3 & 5.6 & 38.0 & 0.17 & 40.2 & 25.3 & 6.4 & 36.8 & 0.17 & 9.5 & 5.4 & 0.7 & 8.7 & 0.05 \\

UnivIntruder~\cite{xu2025one} & 14.1 & 7.4 & 1.3 & 17.9 & 0.05 & 18.6 & 11.3 & 2.2 & 21.1 & 0.05 & 9.4 & 5.1 & 0.8 & 10.9 & 0.03 \\

\rowcolor{highlightcyan}
\textbf{Our tarVRoM-Attack}& \textbf{52.0} & \textbf{34.5} & \textbf{9.9} & \textbf{61.7} & \textbf{0.27} & \textbf{52.6} & \textbf{34.8} & \textbf{9.9} & \textbf{56.7} & \textbf{0.25} & \textbf{14.5} & \textbf{8.7} & \textbf{2.2} & \textbf{15.9} & \textbf{0.07} \\

\hline\hline

\multicolumn{16}{l}{\textcolor{gray!90}{\textit{Performance on Seen Samples (Used for Optimization)}}} \\
AnyAttack~\cite{zhang2025anyattack}& 7.9 & 3.2 & 0.3 & 6.2 & 0.04 & 9.2 & 4.1 & 0.3 & 6.4 & 0.04 & 5.6 & 2.3 & 0.3 & 5.0 & 0.03  \\

\rowcolor{gray10}
M-Attack~\cite{li2025a}& 81.7 & 59.5 & 18.6 & 91.2 & 0.53 & 74.1 & 52.9 & 14.2 & 80.3 & 0.44 & 15.0 & 8.5 & 1.2 & 14.5 & 0.08 \\

FOA-Attack~\cite{jia2025adversarial}& \textbf{84.1} & \textbf{60.9} & \textbf{20.1} & \textbf{93.0} & \textbf{0.57} & \textbf{80.0} & \textbf{57.1} & \textbf{17.5} & \textbf{85.4} & \textbf{0.48} & \textbf{18.6} & 10.5 & 2.1 & 18.0 & 0.10\\

\hdashline
\rowcolor{gray10}
UAP~\cite{moosavi2017universal} & 62.6 & 41.1 & 10.5 & 66.7 & 0.32 & 60.2 & 40.7 & 11.2 & 61.3 & 0.29 & 14.9 & 8.3 & 1.6 & 13.8 & 0.07 \\

UnivIntruder~\cite{xu2025one} & 22.9 & 12.8 & 2.4 & 15.0 & 0.07 & 27.3 & 16.5 & 3.2 & 15.0 & 0.07 & 13.4 & 7.3 & 1.1 & 9.2 & 0.05 \\

\rowcolor{highlightcyan}
\textbf{Our TarVRoM-Attack}& 73.5 & 50.5 & 14.0 & 85.5 & 0.39 & 72.7 & 51.4 & 11.3 & 75.5 & 0.36 & 17.8 & \textbf{11.4} & \textbf{2.9} & \textbf{25.0} & \textbf{0.13} \\

\hline\thickhline

\end{tabular}%
}
}

%% file: tables/tbl3_ablation_mca.tex
\vspace{-2mm}
\scriptsize{
\resizebox{\linewidth}{!}{
\setlength\tabcolsep{4pt} 
\renewcommand\arraystretch{1.1}
\begin{tabular}{ccc||cc|cc|cc}
\hline\thickhline


\rowcolor{gray20}
\multicolumn{3}{c||}{\textbf{Components}} & \multicolumn{2}{c|}{\textbf{GPT-4o}} & \multicolumn{2}{c|}{\textbf{Gemini}} & \multicolumn{2}{c}{\textbf{Claude}} \\
\cline{4-9}

\rowcolor{gray20}
TVA & AFV & TR & ASR & AvgSim & ASR & AvgSim & ASR & AvgSim \\
\hline\hline

\multicolumn{9}{l}{\textcolor{gray!90}{\textit{Performance on Unseen Test Samples}}} \\

 & & & 38.0 & 0.17 & 36.8 & 0.17 & 8.7 & 0.05 \\

\rowcolor{gray10}
$\usym{2713}$ & & & 46.7 & 0.20 & 44.7 & 0.19 & 11.3 & 0.06 \\

 & & $\usym{2713}$ & 46.3&0.21&38.7&0.18&10.3&0.06\\

\rowcolor{gray10}
$\usym{2713}$ & $\usym{2713}$ & & 51.0 & 0.22 & 48.0 & 0.21 & 11.7 & 0.07 \\

$\usym{2713}$ & $\usym{2713}$ & $\usym{2713}$ & 52.0 & 0.22 & 49.0 & 0.21 & 10.0 & 0.06 \\

\hline\hline

\multicolumn{9}{l}{\textcolor{gray!90}{\textit{Performance on Seen Samples (Used for Optimization)}}} \\
 & & & 66.7 & 0.32 & 61.3 & 0.29 & 13.8 & 0.07 \\

\rowcolor{gray10}
$\usym{2713}$ & & & 68.5 & 0.33 & 61.0 & 0.28 & 19.5 & 0.10 \\

& & $\usym{2713}$ &73.0&0.34&64.0&0.29& 26.5&0.12\\

\rowcolor{gray10}
$\usym{2713}$ & $\usym{2713}$ & & 78.0 & 0.40 & 68.5 & 0.33 & 25.0 & 0.11 \\

\usym{2713} & \usym{2713} & \usym{2713} & 80.5 & 0.38 & 70.0 & 0.33 & 24.5 & 0.12 \\

\hline\thickhline
\end{tabular}}}

%% file: tables/tbl4_crops_m.tex
\vspace{-2mm}
\scriptsize{
\resizebox{\linewidth}{!}{
\setlength\tabcolsep{5pt}
\renewcommand\arraystretch{1.1}
\begin{tabular}{c||cc|cc|cc}
\hline\thickhline


\rowcolor{gray20}
 & \multicolumn{2}{c|}{\textbf{GPT-4o}} & \multicolumn{2}{c|}{\textbf{Gemini}} & \multicolumn{2}{c}{\textbf{Claude}} \\
\cline{2-7}

\rowcolor{gray20}
\multirow{-2}{*}{\textbf{$m$}} & ASR & AvgSim & ASR & AvgSim & ASR & AvgSim \\
\hline\hline

\multicolumn{7}{l}{\textcolor{gray!90}{\textit{Performance on Unseen Test Samples}}} \\

$m=2$  & 42.0 & 0.19 & 39.3 & 0.17 & 9.0  & 0.05 \\

\rowcolor{gray10}
$m=4$  & 61.7 & 0.27 & 56.7 & 0.25 & 15.9 & 0.07 \\

$m=8$  & \textbf{64.7} & \textbf{0.28} & 60.3 & \textbf{0.26} & 14.3 & 0.07 \\

\rowcolor{gray10}
$m=16$ & 61.7 & 0.27 & \textbf{60.7} & \textbf{0.26} & \textbf{17.3} & \textbf{0.08} \\

\hline\hline

\multicolumn{7}{l}{\textcolor{gray!90}{\textit{Performance on Seen Samples (Used for Optimization)}}} \\
$m=2$  & 69.0 & 0.31 & 57.0 & 0.25 & 23.5 & 0.09 \\

\rowcolor{gray10}
$m=4$  & 85.5 & 0.39 & 75.5 & 0.36 & 25.0 & 0.13 \\

$m=8$  & \textbf{88.0} & \textbf{0.43} & \textbf{81.0} & \textbf{0.39} & 28.5 & 0.13 \\

\rowcolor{gray10}
$m=16$ & 84.0 & 0.41 & 80.5 & 0.38 & \textbf{31.0} & \textbf{0.15} \\

\hline\thickhline
\end{tabular}}}


%% file: tables/tbl5_fast_adaptation.tex
\vspace{-2mm}
\scriptsize{
\resizebox{\linewidth}{!}{
\setlength\tabcolsep{5pt}
\renewcommand\arraystretch{1.1}
\begin{tabular}{c||cc|cc|cc}
\hline\thickhline

\rowcolor{gray20}
 & \multicolumn{2}{c|}{\textbf{GPT-4o}} & \multicolumn{2}{c|}{\textbf{Gemini}} & \multicolumn{2}{c}{\textbf{Claude}} \\
\cline{2-7}

\rowcolor{gray20}
\multirow{-2}{*}{Stage-2 Epoch} & ASR & AvgSim & ASR & AvgSim & ASR & AvgSim \\
\hline\hline

\multicolumn{7}{l}{\textcolor{gray!90}{\textit{With Meta-Init: Unseen Test Samples}}} \\

50  & 54.7 & 0.23 & 49.0 & 0.21 & 11.8 & 0.06 \\
\rowcolor{gray10}
100 & 56.3 & 0.25 & 53.3 & 0.22 & 12.5 & 0.06 \\
200 & 59.7 & 0.26 & \textbf{56.7} & \textbf{0.25} & 11.7 & \textbf{0.07} \\

\rowcolor{gray10}
300 & \textbf{61.7} & \textbf{0.27} & \textbf{56.7} & \textbf{0.25} & \textbf{15.0} & \textbf{0.07} \\

\hline

\multicolumn{7}{l}{\textcolor{gray!90}{\textit{With Meta-Init: Seen Samples (Used for Optimization)}}} \\

50  & 77.5 & 0.36 & 70.0 & 0.31 & 24.0 & 0.12 \\
\rowcolor{gray10}
100 & 83.0 & 0.39 & 72.5 & 0.33 & 24.5 & 0.12 \\
200 & \textbf{87.0} & \textbf{0.40} & \textbf{76.5} & \textbf{0.36} & 22.0 & 0.11 \\

\rowcolor{gray10}
300 & 85.5 & 0.39 & 75.5 & \textbf{0.36} & \textbf{25.0} & \textbf{0.13} \\

\hline\hline

\multicolumn{7}{l}{\textcolor{gray!90}{\textit{w/o Meta-Init: Unseen Test Samples}}} \\

50  & 25.0 & 0.11 & 22.0 & 0.11 & 4.3  & 0.03 \\
\rowcolor{gray10}
100 & 42.3 & 0.18 & 39.0 & 0.17 & 9.3  & \textbf{0.06} \\
200 & 48.0 & 0.21 & 44.0 & 0.19 & 9.7  & \textbf{0.06} \\
\rowcolor{gray10}
300 & \textbf{52.0} & \textbf{0.22} & \textbf{49.0} & \textbf{0.21} & \textbf{10.0} & \textbf{0.06} \\

\hline

\multicolumn{7}{l}{\textcolor{gray!90}{\textit{w/o Meta-Init: Seen Samples (Used for Optimization)}}} \\

50  & 45.5 & 0.23 & 40.5 & 0.19 & 16.0 & 0.08 \\
\rowcolor{gray10}
100 & 68.0 & 0.31 & 58.5 & 0.26 & 24.0 & 0.11 \\
200 & 77.5 & 0.35 & 69.0 & 0.31 & 24.0 & \textbf{0.13} \\
\rowcolor{gray10}
300 & \textbf{80.5} & \textbf{0.38} & \textbf{70.0} & \textbf{0.33} & \textbf{24.5} & 0.12 \\

\hline\thickhline
\end{tabular}}}

    